\documentclass[conference]{IEEEtran}
\usepackage{amsmath,graphicx}
\usepackage[dvipsnames]{xcolor}
\usepackage{amssymb}
\usepackage{bbm}
\usepackage{algorithm}
\usepackage{algorithmic}
\usepackage[inline]{enumitem}
\usepackage{tikz}
\usepackage{adjustbox} 
\usepackage[nolist]{acronym}
\usepackage{subcaption}
\usepackage{roboto}
\usepackage[T1]{fontenc}
\usepackage[normalem]{ulem}
\usepackage{cite}
\usepackage{hyperref}
\usepackage{cleveref}

\acrodef{ai}[AI]{Artificial Intelligence}
\acrodef{ml}[ML]{Machine Learning}
\acrodef{dl}[DL]{Deep Learning}
\acrodef{nn}[NN]{Neural Network}

\def\x{{\mathbf x}}
\def\z{{\mathbf z}}

\def\y{{\mathbf y}}

\def\a{{\mathbf a}}

\def\R{{\mathbb R}}
\def\A{{\mathbb A}}

\def\X{{\mathcal X}}
\def\Z{{\mathcal Z}}
\def\Y{{\mathcal Y}}
\def\S{{\mathcal S}}
\def\C{{\mathcal C}}

\def \encw{{\theta}}
\def \decw{{\gamma}}
\def \simm{{\psi}}
\def \data{{\x}}
\def \dspace{{\X}}
\def \latent{{\z}}
\def \lspace{{\Z}}
\def \action{{\y}}
\def \aspace{{\Y}}
\def \enc{{E}}
\def \dec{{D}}
\def \rr{{R}}

\newcommand{\se}[2]{\mathbf{#1}}
\newcommand{\seanchor}[2]{\mathbf{#1}^{(#2)}}
\newcommand{\ab}[3]{\se{#1}{#2}_{#3}}
\newcommand{\abanchor}[3]{\seanchor{#1}{#2}_{#3}}
\newcommand{\re}[3]{\mathbf{r}_{#3}}

\newcommand{\cossim}[2]{\frac{\inner{#1}{#2}}{\norm{#1}\norm{#2}}}
\newcommand{\distsim}[2]{\norm{#1-#2}}

\newcommand{\eq}{T}

\newcommand{\eqcritic}{g}
\newcommand{\eqcriticbig}{G}

\DeclareMathOperator*{\argmin}{arg\,min}
\newcommand{\inner}[2]{\left\langle \mathbf{#1},\mathbf{#2}\right\rangle}
\newcommand{\norm}[1]{\left\|#1\right\|}
\newcommand{\parenthesis}[1]{\left(#1\right)}
\newcommand{\squareb}[1]{\left[#1\right]}
\newcommand{\curlyb}[1]{\left\{#1\right\}}

\newcommand{\quotes}[1]{`#1'}

\usetikzlibrary{positioning, shapes, shapes.multipart, fit}
\tikzstyle{block} = [rectangle,rounded corners, minimum width = 1cm, minimum height=1cm,text centered,inner sep=0.5mm] 
\tikzstyle{mytext} = [rectangle,rounded corners, minimum width = 0cm, minimum height=0cm,text centered] 
\newcommand{\yshift}{1}

\newcommand{\xshift}{0.5}
\newcommand{\xxshift}{\xshift*2}
\newcommand{\thicknes}{very thick}
\newcommand{\figfont}{lmss}

\crefformat{equation}{(#2#1#3)}
\crefname{equation}{}{}
\newcommand{\titleheader}{This work has been accepted for publication in 2024 IEEE Globecom Workshops (GC Workshops): From bits to semantics: data factorization, encoding, transmission, and privacy protection.}
\makeatletter
\def\ps@IEEEtitlepagestyle{
\def\@oddhead{\mbox{}\scriptsize \titleheader \rightmark \hfil }
}

\title{Relative Representations of Latent Spaces enable Efficient Semantic Channel Equalization}

\author{\IEEEauthorblockN{Tomás Hüttebräucker$^{1}$, Simone Fiorellino$^{2,3}$, Mohamed Sana$^{1}$, Paolo Di Lorenzo$^{2,3}$, Emilio Calvanese Strinati$^{1}$}

\IEEEauthorblockA{$^1$CEA-Leti, Université Grenoble Alpes, F-38000 Grenoble, France\\
$^2$DIET department, Sapienza University, Rome, Italy. $^3$CNIT, Parma, Italy.\\
Email : \{tomas.huttebraucker; mohamed.sana; emilio.calvanese-strinati\}@cea.fr \\
\{simone.fiorellino; paolo.dilorenzo\}@uniroma1.t}}

\begin{document}
\maketitle
\begin{abstract}
In multi-user semantic communication, language mismatche poses a significant challenge when independently trained agents interact. We present a novel semantic equalization algorithm that enables communication between agents with different languages without additional retraining. Our algorithm is based on relative representations, a framework that enables different agents employing different neural network models to have unified representation. It proceeds by projecting the latent vectors of different models into a common space defined relative to a set of data samples called \textit{anchors}, whose number equals the dimension of the resulting space. A communication between different agents translates to a communication of semantic symbols sampled from this relative space. This approach, in addition to aligning the semantic representations of different agents, allows compressing the amount of information being exchanged, by appropriately selecting the number of anchors. Eventually, we introduce a novel anchor selection strategy, which advantageously determines prototypical anchors, capturing the most relevant information for the downstream task. Our numerical results show the effectiveness of the proposed approach allowing seamless communication between agents with radically different models, including differences in terms of neural network architecture and datasets used for initial training. 
\end{abstract}

\section{Introduction}
\footnotetext{The present work was supported by the EU Horizon 2020 Marie Skłodowska-Curie ITN Greenedge (GA No. 953775), by ``6G-GOALS", an EU-funded project, and by the French project funded by the program \quotes{PEPR Networks of the Future} of France 2030.}
Traditional communication systems are designed to solve the fundamental problem of communication: reproducing at one point either exactly of approximately a message selected at another point \cite{shannon1950mathematical}. Based on this, communication systems are developed to be agnostic to the underlying meaning of the data or how this meaning will affect performance. While for the past decades this approach has found great success at enabling high data transmission rates with low bit-error probabilities, the recent advances in \ac{ai}, particularly \ac{dl}, call for a paradigm shift. Indeed, in \ac{ai} enhanced networks, intelligent agents based on \acp{nn} exchange the learned representation of the data rather than the data itself, which essentially constitutes an exchange of its meaning (semantics). Moreover, \ac{nn} are robust to noise in the input, which means that designing \ac{ai} based communication systems as opaque data pipelines optimized solely for low bit-error rates is an inefficient use of valuable resources. Semantic communications have recently emerged a new communication framework where the focus is put on correctly conveying the meaning of the data to enable the completion of the underlying task. Employing \ac{dl} to design semantic communication protocols, also referred to as semantic languages, has been shown to largely outperform traditional communication systems both in classification \cite{bourtsoulatze2019deep} and sequential control problems \cite{tung2021effective}. So far, most works in the semantic communications literature share one key assumption: the communication language has been previously learned and is fixed during deployment. This is a natural choice, as it has been observed in the \ac{ml} literature that independently learned languages can vary significantly, even when the architecture, data, and optimization objectives are identical \cite{moschella2023relative}. However, assuming that all agents participating in the communication have undergone a joint training procedure limits the application of semantic communications in dynamic environments. In such scenarios, like autonomous vehicles or IoT networks, the agents and their objectives must continuously adapt to changing conditions. Regularly unifying the language is an energy consuming process that could outweigh the advantages of semantic communications. Therefore, it is crucial to ensure that the semantic protocols in use enable effective task-solving under such dynamic scenarios. 

The divergence in semantic languages, referred to as semantic mismatch, was first studied in \cite{sana2022learning} and addressed in \cite{sana2023semantic}, where the authors proposed the Semantic Channel Equalization algorithm to \textit{equalize} the semantic mismatch between a transmitter and a receiver. This algorithm relies on a codebook of low-complexity linear transformations between the \textit{atoms} (regions of shared semantic meaning) of the transmitter and receiver latent spaces which is operated by a selection policy. While this approach has been shown to be effective, it requires identifying and defining the atoms of different models, which is not straightforward. In \cite{huttebraucker2024soft}, this challenge was addressed by defining the atoms in a self-supervised manner, leading to improved results. However, the atom identification process still requires sharing data, and the transformation selection policy depends on atom estimation, which remains a complex problem. In this work, we propose a novel approach that leverages the concept of \textit{relative representations} \cite{moschella2023relative} to establish a common communication channel between different semantic protocols in a zero-shot manner, without the need for retraining the models or labeled data. Relative representations were introduced as a method to project the data embeddings of different models into a unified representation space. Instead of relying on the latent coordinates of each model, the authors propose to use a relative representation whose components are the similarity score between the desired data embedding and a set of predefined points called \textit{anchors}. This way, the semantic meaning extracted by different models could be unified if the similarity function is sensible to the way these models encode information. The effectiveness of relative representations to unify representations has been shown for text and image data \cite{moschella2023relative} and for reinforcement learning problems \cite{ricciardi2023zero}. Also, multiple metrics to compute the similarity score have been explored \cite{cannistraci2023bricks}, and the generalization capabilities have been shown \cite{norelli2023asif}. Moreover, \cite{fiorellino2024dynamic} introduces an optimization strategy that leverages relative representations to dynamically optimize communication, computation, and learning resources, such as anchor sets and encoders. This approach aims to enable energy-efficient, low-latency, and effective semantic communications. All these works, however, assume it is possible to train a relative decoder in the new representation space, which might be infeasible in resource-constrained networks. In contrast, in \cite{maiorca2024latent} the authors leverage the cosine similarity based relative representations that admits an inverse to perform alignment between models without the need to train a new decoder. In this work, we follow the same principle, using the relative space as a common communication channel to enable translation between different models. Our main contributions are: 
\begin{enumerate*}
    \item We propose a semantic equalization framework based on relative representations that does not need any retraining and requires sharing a small amount of data between transmitter and receiver. Different from \cite{maiorca2024latent}, our method is agnostic to the similarity function of relative representations.
    \item We introduce a novel anchor selection algorithm called prototypical anchors, which improves the performance of semantic equalization compared to classical approaches such as random selection.
\end{enumerate*}

\section{System Model}

\begin{figure*}[ht!]
    \fontfamily{\figfont}
    \centering
    \begin{subfigure}{0.95\textwidth}
        \centering
        \begin{adjustbox}{width=0.98\textwidth,center}
        \input{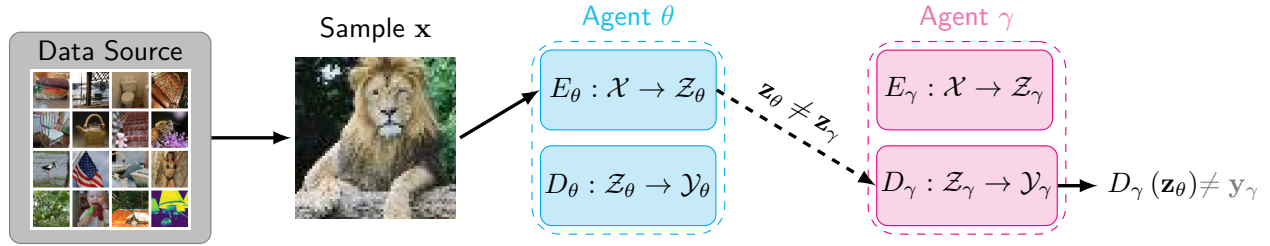}
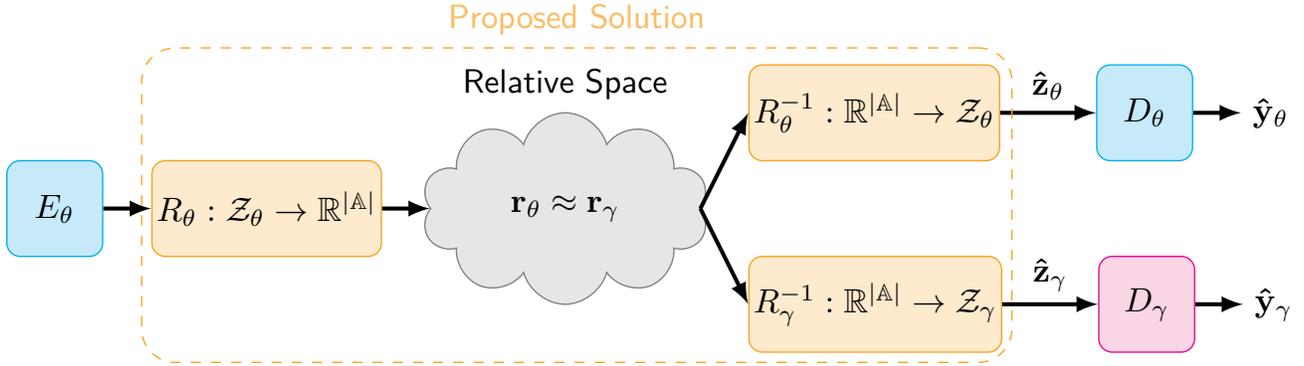
        \end{adjustbox}
        \caption{Multi-user semantic mismatch: Two independently trained models ($\encw, \decw$) represent the same data differently. Language mismatch arises when pairing $E_\encw$ with $D_\decw$.}
        \label{fig:system_model}
    \end{subfigure}

    \vspace{0.5cm}  

    \begin{subfigure}{0.96\textwidth}
        \centering
        \begin{adjustbox}{width=1\textwidth,center}
        \begin{tikzpicture}[node distance= 0cm]

    \node (semencoder21) [block, fill=Cyan!20, draw=Cyan, yshift=0 cm] at (-0.75,-4) {$\enc_\encw$};

    {\fontfamily{cmr}

    \node (relencoder21) [block, fill=YellowOrange!20, draw=YellowOrange, right=\xshift cm of semencoder21] {$\rr_\encw:\lspace_\encw\to\R^{|\A|}$};
    \draw [-latex,\thicknes] (semencoder21.east)-- (relencoder21.west) node[midway,above]{};
    }

    \node (relativespace) [cloud, cloud puffs=10, cloud ignores aspect, minimum height=2cm, minimum width=3cm, draw=gray, fill=gray!20, right=\xshift cm of relencoder21, yshift=0 cm] {$\re{\latent}{i}{\encw}\approx\re{\latent}{i}{\decw}$};
    \node (relativespace_text) [mytext,above=0.01cm of relativespace]{Relative Space};
    \draw [-latex,\thicknes] (relencoder21.east)-- (relativespace.west) node[midway,above]{};

    {\fontfamily{cmr}
    \node (reldecoder21) [block, fill=YellowOrange!20, draw=YellowOrange, right=\xshift cm of relativespace, yshift=\yshift cm] {$\rr_\encw^{-1}:\R^{|\A|}\to\lspace_\encw$};
    \draw [-latex,\thicknes] (relativespace.east)-- (reldecoder21.west) node[midway,above]{};
    }
    
    \node (semdecoder21) [block,fill=Cyan!20, draw=Cyan, right=\xxshift cm of reldecoder21]{$\dec_\encw$};
    \draw [-latex,\thicknes] (reldecoder21.east)-- (semdecoder21.west) node[midway,above,]{$\ab{\hat{\latent}}{i}{\encw}$};

    \node(action_semantic21) [mytext, right=\xshift cm of semdecoder21] {$\ab{\hat{\action}}{i}{\encw}$};
    \draw [-latex,\thicknes] (semdecoder21.east)-- (action_semantic21) {};

    {\fontfamily{cmr}
    \node (reldecoder22) [block, fill=YellowOrange!20, draw=YellowOrange, right=\xshift cm of relativespace, yshift=-\yshift cm] {$\rr_\decw^{-1}:\R^{|\A|}\to\lspace_\decw$};
    \draw [-latex,\thicknes] (relativespace.east)-- (reldecoder22.west) node[midway,above]{};
    }
    
    \node (semdecoder22) [block,fill=Magenta!20, draw=Magenta, right=\xxshift cm of reldecoder22]{$\dec_\decw$};
    \draw [-latex,\thicknes] (reldecoder22.east)-- (semdecoder22.west) node[midway,above,]{$\ab{\hat{\latent}}{i}{\decw}$};

    \node(action_semantic22) [mytext, right=\xshift cm of semdecoder22] {$\ab{\hat{\action}}{i}{\decw}$};
    \draw [-latex,\thicknes] (semdecoder22.east)-- (action_semantic22) {};

    \node (box) [draw=YellowOrange,rounded corners=.35cm, dashed, inner sep=0.1cm, fit=(reldecoder21)(relencoder21) (relativespace_text)(reldecoder22)] {};
    \node [below, text=YellowOrange!70, above=0.01 cm of box] {Proposed Solution};

\end{tikzpicture}
        \end{adjustbox}
        \caption{Proposed solution leveraging relative representations: The relative encoders $R$ project semantic messages into the shared relative space. The relative decoders $R^{-1}$ recover the semantic message corresponding to each decoder.}
        \label{fig:proposed_solution}
    \end{subfigure}
    
    \caption{System model showing the language mismatch and our proposed solution to create a common communication channel between independently trained models.}
    \label{fig:model_solution}
\end{figure*}

We focus on the multi-user semantic communications scenario depicted in \cref{fig:system_model}. A transmitter (encoder $E$) observes data $\se{\data}{i}\in\dspace$ (e.g, an observation of the environment or a sample from an information source) and sends some features of $\se{\data}{i}$ to a remote receiver (decoder $D$) with the goal of successfully completing a downstream task. In traditional communications, the transmitter would use a compression algorithm together with channel coding to transmit $\se{\data}{i}$ in an error-free fashion, and the decoder would recover $\se{\hat{\data}}{i}$ to perform inference and complete the task. In semantic communications, the encoder transmits the \quotes{meaning} of the data, and the decoder performs inference based on the interpretation of the sent message without recovering the original $\se{\data}{i}$. In practice, the encoder and decoder are modeled by \acp{nn}, which follow a \ac{ml} training process to jointly learn a semantic language. The result of the learning are the parameters $\encw=\curlyb{\encw_\text{enc},\encw_\text{dec}}$ of the encoder $E_{\encw_\text{enc}}$ and decoder $D_{\encw_\text{dec}}$. To simplify the notation, we will just make reference to the learned global parameters $\encw$ to reference the weights of the encoder and the decoder, i.e. $E_{\encw_\text{enc}}=E_\encw$ and $D_{\encw_\text{dec}}=D_\encw$. Using the learned language, the transmitter $E_{\encw}$ extracts the task relevant information of $\se{\data}{i}$ and maps it into a semantic representation $\ab{\latent}{i}{\encw}=E_{\encw}\parenthesis{\se{\data}{i}}\in\lspace_\encw$. On the other end of the communication, the decoder receives the semantic representation $\ab{\latent}{i}{\encw}$ and outputs a decision based on it $\ab{\action}{i}{\encw}=D_{\encw}\parenthesis{\ab{\latent}{i}{\encw}}\in\aspace_\encw$.  

In multi-user communications scenario (as shown in \cref{fig:system_model}), the transmitter and receiver may maintain two different models ($E_\encw$, $D_\encw$) and ($E_\decw$, $D_\decw$) trained on similar tasks (defined by dataset, goal or environment). To enable effective collaboration, the task-relevant data features extracted by $E_\encw$ ($E_\decw$) should be consistently decoded by $D_\decw$ ($D_\encw$). However, even when the task is shared, if the models are trained independently, it is unlikely that the encoders will represent data $\se{\data}{i}$ in a consistent manner, (i.e. $\ab{\latent}{i}{\encw}\neq\ab{\latent}{i}{\decw}$). This is an undesirable and usually unavoidable issue in \acp{nn}. Indeed, even in the absence of stochastic factors during the training procedure, architectural choices such as the structure and the dimensions of the \acp{nn} layers may result in different latent space representations causing defective communication.

\subsection{Semantic Channel Equalization}
\label{subsec:semantic_mismatch}
Semantic Channel Equalization focuses on aligning the semantic representations of independently trained agents to enable heterogeneous communication. A semantic equalizer is a transformation $\eq_{\encw\to\decw}:\lspace_\encw\to\lspace_\decw$ between the transmitter (source) semantic space $\lspace_\encw$ and the receiver (target) semantic space $\lspace_\decw$. The latent equalizer aims to align (in some sense) the source and target representation spaces. In general, the effectiveness of the channel equalizer $\eq_{\encw\to\decw}$ can be measured by 
\begin{equation}
    \eqcriticbig\parenthesis{\eq_{\encw\to\decw}} = \mathbb{E}_{\se{\data}{i}\sim\dspace}\squareb{\eqcritic\parenthesis{\eq_{\encw\to\decw}\parenthesis{\ab{\latent}{i}{\encw}},\ab{\latent}{i}{\decw}}},
\end{equation}
where $\eqcritic:\lspace_\decw\times\lspace_\decw\to\R$ is a measure function (e.g., a distance measure) between the transformed source semantic message $\eq_{\encw\to\decw}\parenthesis{\ab{\latent}{i}{\encw}}$ and the target message $\ab{\latent}{i}{\decw}$. For example, if the goal of the equalizer is the perfect alignment of the languages, the function $\eqcritic$ could be the squared Euclidean distance between the semantic messages $\eqcritic_\text{SE}=-\norm{\eq_{\encw\to\decw}\parenthesis{\ab{\latent}{i}{\encw}}-\ab{\latent}{i}{\decw}}^2$. While this seems like the most natural choice of equalization objective, imperfect alignment will not necessarily lead to a degradation in task performance. Indeed, \acp{nn} are robust to some noise in the input data, which means that some alignment error can be tolerated by the decoder. Thus, an alternative is to measure the alignment quality by comparing the output of the decoder in a goal-oriented manner $\eqcritic_\text{GO}=\mathbbm{1}\parenthesis{D_\decw\parenthesis{\eq_{\encw\to\decw}\parenthesis{\ab{\latent}{i}{\encw}}},D_\decw\parenthesis{\ab{\latent}{i}{\decw}}}$. In previous works \cite{sana2023semantic}\cite{huttebraucker2024soft}, authors focused on developing equalizers that were optimal for $\eqcritic_\text{GO}$. In our work, we will not choose to optimize for any of these metrics explicitly, but we will show how our method performs based on them. 

\subsection{Relative Representations} 
Even if the way two independently trained encoders represent the same set of features extracted from the data can greatly vary, the representation spaces of effective feature extractors share a certain structure \cite{moschella2023relative}. Relative representations leverage this to find a common projection space based on how a shared set of data points named \textit{anchors} are represented by each encoder. Starting from a (a priori randomly selected) subset of the data space $\A=\curlyb{\seanchor{\a}{1},\seanchor{a}{2},\ldots,\seanchor{a}{|\A|}}\subset\dspace$ called the \textit{anchor set}, the relative representation of a semantic message $\ab{\latent}{i}{\encw}=E_\encw\parenthesis{\se{\data}{i}}$ is computed as 
\begin{equation}
\label{eq:relrep}
    R_{\A_\encw}^{\simm}\parenthesis{\ab{\latent}{i}{\encw}}=\re{\latent}{i}{\encw} = \squareb{\simm\parenthesis{\ab{\latent}{i}{\encw},\abanchor{\a}{1}{\encw}},\ldots,\simm\parenthesis{\ab{\latent}{i}{\encw},\abanchor{\a}{|\A|}{\encw}}},
\end{equation}
where $\A_\encw=\curlyb{\abanchor{\a}{1}{\encw},\abanchor{\a}{2}{\encw},\ldots,\abanchor{\a}{|\A|}{\encw}}$ is the set of \textit{absolute} semantic representations of the anchors, i.e. $\abanchor{\a}{j}{\encw}=E_\encw\parenthesis{\seanchor{\a}{j}}$, and $\simm:\lspace_\encw\times\lspace_\encw\to\R$ is a \textit{similarity} metric that measures how similar two absolute semantic representations are. To ease the notations, we later refer to $R_{\A_\encw}^{\simm}$ as $R_\encw$, where the dependence of the similarity function $\simm$ and anchor set $\A$ is implicit. The relative projection in \cref{eq:relrep} transforms the absolute representation $\ab{\latent}{i}{\encw}\in\lspace_\encw$ into a relative one $\re{\latent}{i}{\encw}\in \R^{|\A|}$. Each dimension of $\re{\latent}{i}{\encw}$ captures how \quotes{similar} $\ab{\latent}{i}{\encw}$ is to each of the anchors in $\A_\encw$.

As shown in \cite{moschella2023relative,ricciardi2023zero,cannistraci2023bricks,norelli2023asif,fiorellino2024dynamic,maiorca2024latent}, for certain common similarity functions and well-behaved encoders, the relative projection has two properties that make it well-suited to unify different semantic spaces. First, for two encoders $E_\encw$ and $E_\decw$, the relative representations of the same data samples $x$ (using the same similarity function and set of anchors) are similar
\begin{equation}
    \re{\latent}{i}{\encw} \approx \re{\latent}{i}{\decw}, \forall\; \se{\data}{i}\in\X.
\end{equation}
Second, the relative representations also carry the task-relevant information: a decoder $D_\text{rel}$ trained on the relative representation of an encoder $E_\encw$ can successfully learn the given task with similar performance as a decoder $D_\encw$ trained using the absolute representations of $E_\encw$.

This way, relative representations serve as a common semantic channel between multiple encoders. Each encoder can transmit its own relative representation of data, and it will be consistent with any other relative representation of different encoders. Furthermore, since the dimension of the relative representation depends on the cardinality of the anchor set, $\re{\latent}{i}{\encw}\in\R^{|\A|}$, this framework enables information compression. Indeed, depending on the channel state and decoder capabilities, the encoder may choose to use a larger or smaller number of anchors, consuming more or less resources and enabling a dynamic optimization of the network resources \cite{fiorellino2024dynamic}.

\section{Proposed Solution}
We aim to design the equalizer $\eq_{\encw\to\decw}$ leveraging the common communication channel enabled by the relative representations. If the relative projection was an invertible process, it would be possible for a decoder $D_\decw$ to recover a semantic representation from the relative representation $\re{\latent}{i}{\encw}$ transmitted by a non-matching encoder $E_\encw$. The  communication process through the common relative channel follows as
\begin{align}
\label{eq:communication_pipeline}
     \se{\data}{i} \xrightarrow[]{E_\encw} \ab{\latent}{i}{\encw} \xrightarrow[]{R_\encw} \re{\latent}{i}{\encw} \approx 
    \re{\latent}{i}{\decw} \xrightarrow[]{R_\decw^{-1}} \ab{\hat{\latent}}{i}{\decw} \xrightarrow[]{D_\decw} \se{\hat{y}}{i}.
\end{align}
Where $R_\encw$ and $R_\decw^{-1}$ are the transmitter relative projection (shown in \cref{eq:relrep}) and receiver \textit{relative projection inverse} functions respectively. In order to successfully implement the communication pipeline shown in \cref{eq:communication_pipeline}, the following conditions have to be met:

\noindent\textbf{Relative Representation Equalizer Conditions}
\begin{enumerate}[label=(\roman*)]
\item \label{enum:rel_similarity} The relative space is encoder invariant, i.e.,
\begin{equation}
    \re{\latent}{i}{\encw} \approx \re{\latent}{i}{\decw}
\end{equation}
\item \label{enum:rel_inverse}It is possible to pseudo-invert the relative projection, i.e.,
\begin{align}
\begin{split}
&\forall \re{\latent}{i}{\encw} \in \R^{|\A|}, \textrm{ } \exists \textrm{ } \ab{\hat{\latent}}{i}{\decw} \in\lspace_\decw, \textrm{ such that }\\
&R_\decw\parenthesis{\ab{\hat{\latent}}{i}{\decw}} = \re{\latent}{i}{\encw} \approx \re{\latent}{i}{\decw} \textrm{ and }\\
&R_\decw^{-1}\parenthesis{\re{\latent}{i}{\encw}}=\ab{\hat{\latent}}{i}{\decw}\\
\end{split}
\end{align}
\item \label{enum:rel_info} The relative representation $\re{\latent}{i}{\decw}$ carries the information about $\ab{\action}{i}{\decw}$ (a relative decoder effectively solves the task). This enables
\begin{equation}
    D_\decw(\ab{\hat{\latent}}{i}{\decw})\approx D_\decw(\ab{\latent}{i}{\decw})
\end{equation}
\end{enumerate}
Our proposed latent equalizer can then be fully described as 
\begin{equation}
    \eq_{\encw\to\decw}=R_\decw^{-1}\circ R_\encw.
\end{equation}
In the following sections, we will introduce the design of the relative inversion function $R_\decw^{-1}$ assuming that condition \ref{enum:rel_inverse} is met. Furthermore, we introduce an anchor selection algorithm that enhances the fulfillment of conditions \ref{enum:rel_similarity} and \ref{enum:rel_info}, enhancing the performance of the system.

\subsection{Relative Projection Inverse $R_\decw^{-1}$}

The relative representation inverse is a function $R_\decw^{-1}:\R^{|\A|}\to\lspace_\decw$ that recovers the absolute representation from the received relative leveraging the set of pre-selected anchors of the decoder. The existence of the inverse of the relative projection depends on the anchor set $\A$ and the similarity function $\simm$. For example, if $\simm$ is chosen to be the well-known cosine similarity function, it is impossible to find the exact inverse since only the angle information of the absolute space is preserved. On the other hand, if the selected $\simm$ is the Euclidean distance, it will be possible to find an exact inverse if the number of anchors is higher than the dimension of the latent space. Nevertheless, for effective task solving, it is not necessary to find the exact inverse, but rather the pseudo-inverse as described in condition \ref{enum:rel_inverse}. This can be done by solving the following optimization problem
\begin{equation}\label{eq:inverse}
    \ab{\hat{\latent}}{i}{\decw} = \argmin_{\latent_\decw\in\lspace_\decw}\norm{R_{\decw}\parenthesis{\latent_\decw}-\re{\latent}{i}{\encw}}^2
\end{equation}
In this work, we choose to solve the optimization problem in \cref{eq:inverse} by using a gradient descent-based optimization algorithm. A pseudo algorithm for our method is shown in \cref{alg:rel_inverse}.

\begin{algorithm}[h]
\caption{\strut Relative Representation Inverse $R_\decw^{-1}$}
\label{alg:rel_inverse} 
\begin{algorithmic}[1]
\REQUIRE $\re{\latent}{i}{\encw},\A_\decw, \text{max\_it}$
\ENSURE $\ab{\hat{\latent}}{i}{\decw}$
\STATE Randomly choose initial guess $\latent_1\sim\text{Uniform}\parenthesis{\lspace_\gamma}$;
\FOR {$i \gets 1$ to $\text{max\_it}$}
    \STATE Compute squared error  $L=\norm{\re{\latent}{i}{\encw}-R_{\decw}\parenthesis{\latent_i}}^2$
    \STATE Apply optimization algorithm using $\frac{\partial L}{\partial \latent_i}$ and obtain $\latent_{i+1}$ 
\ENDFOR
\RETURN $z_{\text{max\_it}}$
\end{algorithmic}
\end{algorithm}

Notice that the problem in \cref{eq:inverse} can be solved in closed form depending on the similarity function $\simm$. For example, in \cite{maiorca2024latent}, the authors use the cosine similarity and obtain the inverse as a closed-form solution. Our method has the advantage of being agnostic to the similarity function $\simm$.

\subsection{Prototypical Anchors}
\label{sec:alignment2}

The anchor set $\A$ is a key aspect of the relative representation. The information carried by the relative projection, how well different relative representations align, and the resources used in the communication all depend on it. Ideally, $\A$ should enable the transmission of the task-relevant information and allow a high level of alignment in the relative space while being composed of the least number of elements possible. To accomplish this, each anchor in $\A$ should be representative of a distinct task-relevant feature of the data to be encoded in the relative vector. However, selecting the samples that better represents the task relevant features is not straightforward. To overcome this problem, we propose to exploit the fact that the absolute space of a well-performing encoder represents the features in a structured manner. Samples in such a latent space are mapped according to the features they contain, and samples with similar features should be mapped\quotes{closely}. Effectively, all samples that share a feature or a set of features will be mapped into the same region of the space, from which a good anchor candidate can be obtained. To explicitly define these regions, we propose to leverage a self-supervised clustering algorithm that divides the space and whose centroids can be used as anchors. This is the main idea of the prototypical anchors algorithm (shown in \cref{alg:proto_anchors}) whose name is inspired on prototypical neural networks \cite{snell2017prototypical}.

\begin{algorithm}[h]
\caption{Anchor Prototypes} 
\label{alg:proto_anchors} 
\begin{algorithmic}[1]
\REQUIRE $\curlyb{\dspace,N,M}$ or $\S$, $E_\encw$
\ENSURE  Anchor set $\A_\encw$
\IF{$\S$ is not given}
    \STATE Compute dataset absolutes: $Z_\encw \gets E_\encw(\dspace)$
    \STATE Cluster absolute space:  $\C = \{C_1, \ldots, C_{N}\} \gets$ Apply a clustering technique with $N$ clusters to $Z_\encw$, $\bigcup_{i=1}^{N} C_i = Z_\encw$
    \STATE Support set: $\S = \{S_1, \ldots, S_{N}\} \gets$ Sample $M$ elements from each cluster, forming $\{S_1, \ldots, S_{N}\}$, with $S_i \subset C_i$
    \STATE Fix $\S$ as input for all other encoders
\ENDIF
\STATE Compute the anchor set: $\A_\encw = \{\abanchor{a}{1}{\encw}, \ldots , \abanchor{a}{N}{\encw}\} \gets \abanchor{a}{i}{\encw} = \frac{1}{M}\sum_{s\in S_i} E_\encw\parenthesis{s}$
\STATE \textbf{return} $\A_\encw(\S)$
\end{algorithmic}
\end{algorithm}
\noindent We perform clustering on the latent space of a chosen encoder, setting the number of clusters $N$ equal to the desired number of anchors $|\A|$. Ideally, the clustered space should be divided according to the $N$ most relevant semantic features of the data. Each of the cluster centroids should be the best representative of the cluster feature since it maximizes the difference (measured according to the clustering algorithm metric) to other clusters. However, choosing the cluster centroids is not possible since they are not grounded in the data but rather in the absolute space of the selected encoder. This is not translatable to other encoders' latent spaces. For this reason, we estimate the centroids by randomly selecting $M$ samples from each cluster and setting the mean of their absolute representation as the centroid estimation. This way, by sharing the data samples, each encoder can independently compute its anchor set.

\section{Results}
We tested our performance on an image classification task on the tiny-imagenet dataset \cite{tinyimagenetHF}, which is comprised of 100k $64\times64$ colored images of 200 balanced classes. We selected a set of feature extractor models pre-trained on the Imagenet classification challenge \cite{modelsHF} to serve as the encoders of our scenario. For each of the feature extractors, we train a classifier network on top that we use as the decoder. This way, we obtain a set of different encoders with the corresponding decoders. For our proposed relative representation inverse $R_\decw^{-1}$ (\cref{alg:rel_inverse}) we set $\text{max\_it}=1000$ and use the Adam algorithm \cite{kingma2014adam} to perform the gradient based optimization. For the prototypical anchors (\cref{alg:proto_anchors}), we use KMeans as the clustering algorithm, we set $M=5$, and we use the transmitter's encoder as the encoder that fixes the set $\S$. For the similarity metric $\simm$, we explore the cosine similarity,
\begin{equation}
    \text{cosine similarity}\parenthesis{\ab{\latent}{i}{\encw},\abanchor{\a}{j}{\encw}} = \cossim{\ab{\latent}{i}{\encw}}{\abanchor{\a}{j}{\encw}},
\end{equation}
and the normalized Euclidean distance,
\begin{equation}
    \text{distance}\parenthesis{\ab{\latent}{i}{\encw},\abanchor{\a}{j}{\encw}} =\frac{\distsim{\ab{\latent}{i}{\encw}}{\abanchor{\a}{j}{\encw}}}{\frac{1}{|\A|}\sum_{\seanchor{\a}{k}\in\A}\norm{\abanchor{\a}{k}{\encw}}}.
\end{equation}
Both the cosine similarity and Euclidean distance have been widely used in the relative representations literature \cite{moschella2023relative}\cite{cannistraci2023bricks}. Different from previous works where the authors normalize the representation space to project on the relative one, in this work, we aim to recover the absolute representation, and thus, we normalize the Euclidean distance by the mean of the anchor set. This way, we ensure that the distance-based relative representation is invariant to scale differences in the semantic spaces, and we are able to recover $\ab{\latent}{i}{\decw}$ from it.  
We will compare our results with the pseudo-inverse solution proposed in \cite{maiorca2024latent} where they leverage the cosine similarity-based relative representation expression to obtain a closed form solution for \cref{eq:inverse}:
\begin{equation}
    R_\decw^{-1}\parenthesis{\re{\latent}{\simm}{\encw}} = \re{\latent}{\simm}{\encw} \parenthesis{\overline{\A}_\decw^T}^{-1}
\end{equation}
where $\overline{\A_\decw}\in\R^{|\A|\times d_\decw}$ is the row-normalized target anchor matrix, where each row is an anchor target absolute representation $\parenthesis{\lspace_\decw=\R^{d_\decw}}$ and$\parenthesis{\A_\decw^T}^{-1}$ is the row normalized target anchor matrix generalized inverse which can be computed as 
\begin{equation}
    \parenthesis{\A_\decw^T}^{-1} = \parenthesis{\overline{\A}_\decw^{T}\overline{\A}_\decw}^{-1}\overline{\A}_\decw^{T}.
\end{equation}
\begin{figure}
    \centering
    \includegraphics[width=0.96\linewidth]{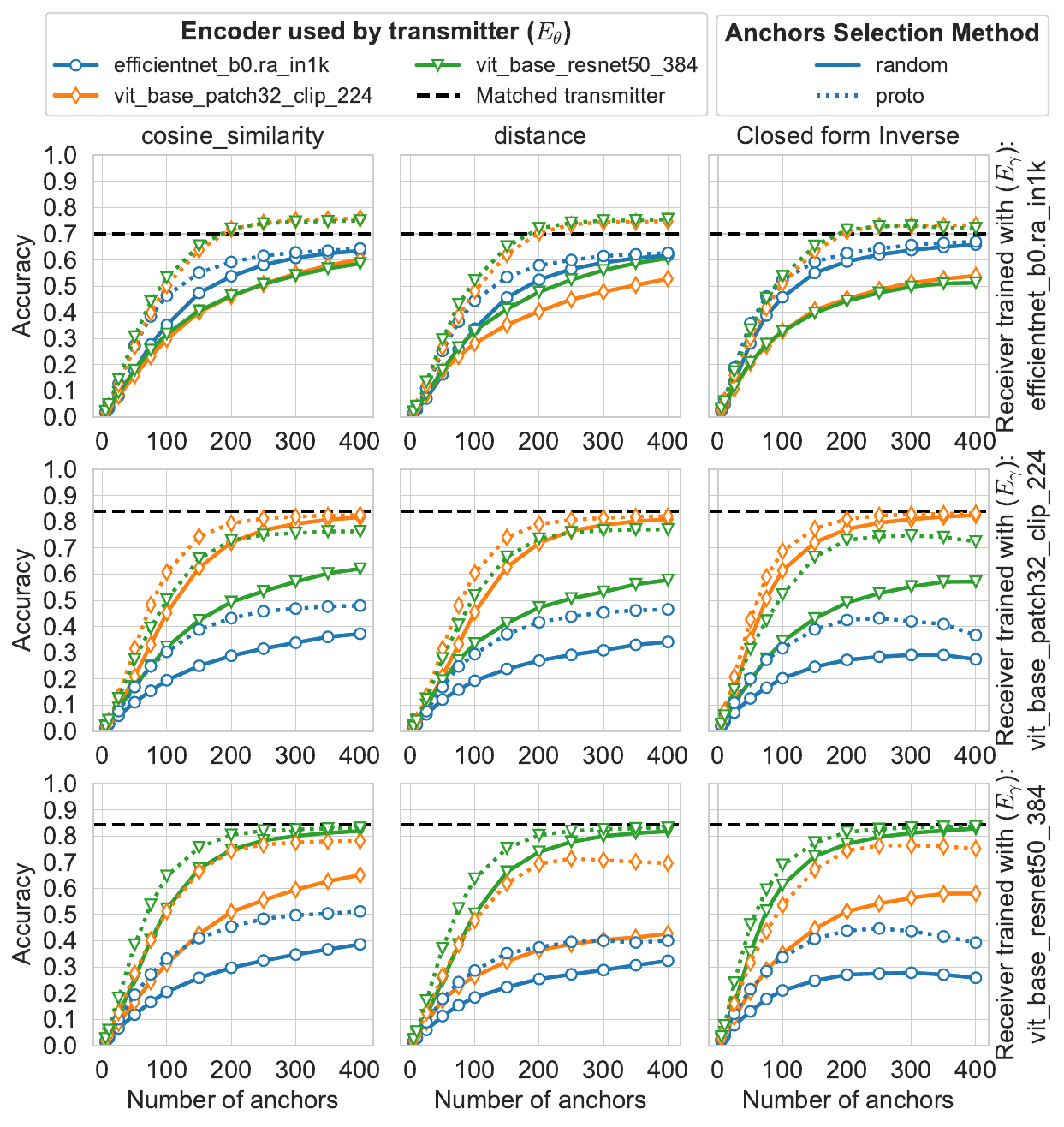}  
    \caption{Performance of the proposed Semantic Channel Equalization algorithm (two leftmost columns) and the closed-form inverse used in \cite{maiorca2024latent} (rightmost column). Results are compared with a random anchor selection and the proposed prototypical anchors.}
    \label{fig:accuracy}
\end{figure}
\begin{figure}
    \centering
    \includegraphics[width=0.96\linewidth]{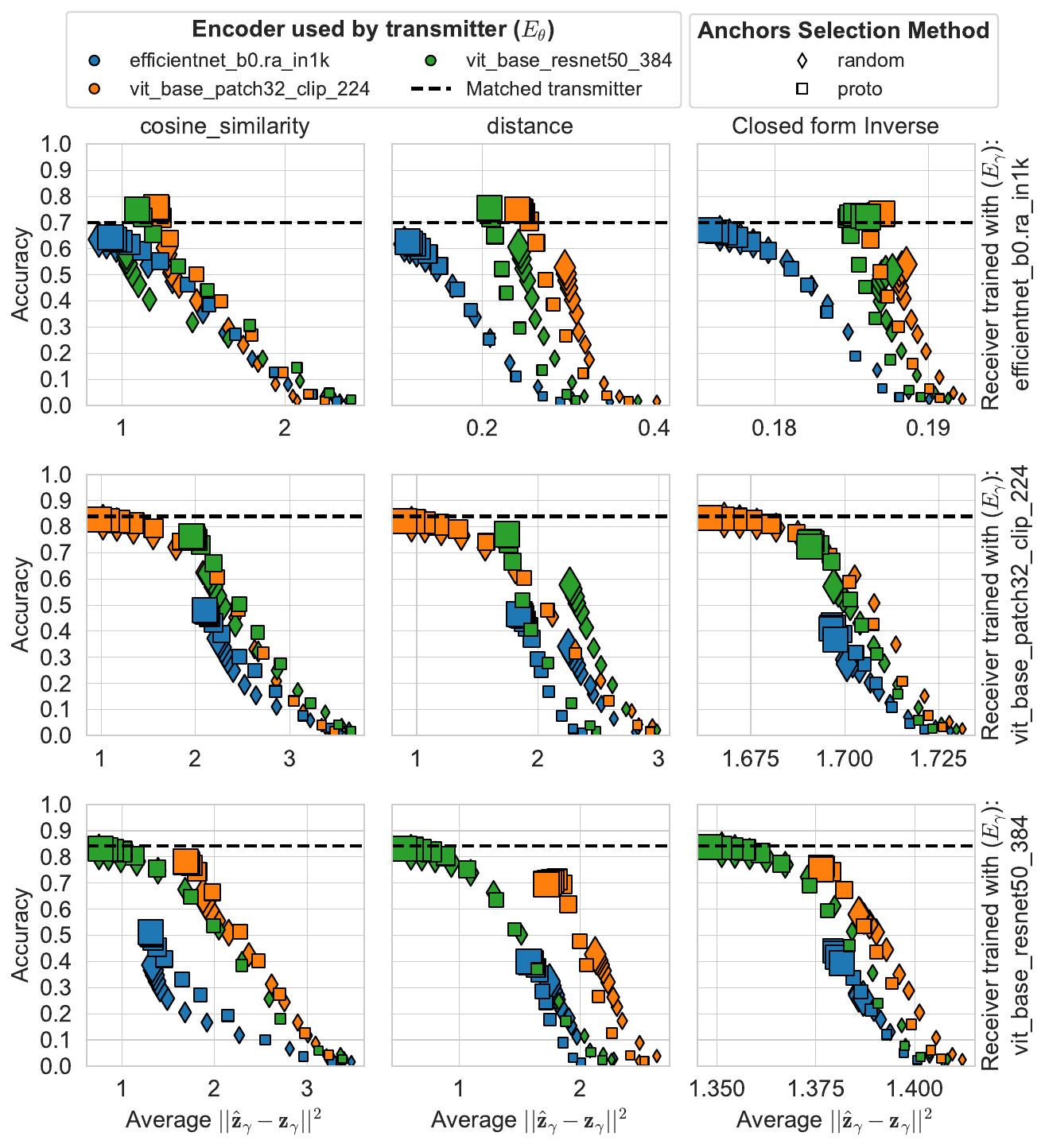}  
    \caption{Accuracy as a function of the error in the reconstruction of the target semantic symbol $\z_\decw$ for the Semantic Channel Equalization method. The number of anchors used is shown in the marker size, bigger marker size corresponds to a higher number of anchors used.}
    \label{fig:accuracy_vs_mse}
\end{figure}
\subsection{Equalization performance}
In \cref{fig:accuracy}, the results of the semantic channel equalization are shown. It is clear that semantic equalization is effective, and as the number of anchors increases, the accuracy does as well. In particular, it is interesting to see that the target decoder $E_\decw$ sometimes performs better when paired with a powerful equalized encoder $E_\encw$ rather than the matched encoder $E_\decw$ (see results when $E_\decw=\text{efficientnet\_b0.ra\_in1k}$). This can be explained by the fact that a more descriptive encoder is capable of transmitting more information and that our method is effective in translating this information into the decoder space. The closed form inverse based equalization has similar performance to our method but looses performance in the high anchor number regime, this can be explained by the instability of the matrix inverse when it's size increases. Moreover, while the closed form method is less complex, it is only applicable to a cosine similarity based relative representation. Our proposed optimization method is agnostic to the similarity metric, and it thus has the potential to perform better with an improved similarity metric. This is of particular interest when the similarity function is learned for a given task. 

Our proposed prototypical anchors method enhances the performance of the equalization algorithms in almost all the experiments. This confirms the intuition behind the design of the prototypical anchors algorithm. Moreover, this algorithm can be further improved by appropriately tuning its parameters, especially the clustering method and the centroid estimation technique used. The choice of the clustering algorithm is crucial since it assumes the way that information is encoded in the absolute space. By choosing KMeans as a clustering algorithm we assume that the encoder uses the Euclidean distance as a way to encode the information. Other metrics can be explored for this but exploring this is out of the scope of the paper.

\subsection{Semantic vs Goal oriented equalization performance}
In \cref{subsec:semantic_mismatch} we introduced two ways to measure the equalization performance $\eqcritic_\text{GO}$ and $\eqcritic_\text{SE}$. While $\eqcritic_\text{GO}$ evaluated the effect of the equalization as how it affects the task performance, $\eqcritic_\text{SE}$ compared the output of the equalization algorithm $\hat{\latent}_\decw$ with the encoder matched output $\latent_\decw$. In \cref{fig:accuracy_vs_mse} we show the accuracy (related to $\eqcritic_\text{GO}$) as a function of the reconstruction error $\norm{\hat{\latent}_\decw-\latent_\decw}^2$ (equal to $\eqcritic_\text{SE}$). We see that there is a clear correlation between both: as the reconstruction error decreases, the accuracy increases. However, it is not always true that smaller reconstruction errors lead to increased accuracy. In many cases, better performance is obtained with a higher reconstruction error, which shows one key characteristic of semantic communications and \acp{nn}.

\section{Conclusions and Future Work}

In this work, we introduced a Semantic Channel Equalization algorithm based on relative representations, enabling a receiver to decode messages from an independently trained transmitter by exchanging only a small set of data points known as anchors. The algorithm supports compression, with data size depending on the number of anchors; including more anchors improves performance but reduces the compression ratio. We present a novel anchor selection strategy called prototypical anchors, which partitions the semantic space of an encoder to capture essential data features. Our results show that using prototypical anchors significantly enhances semantic equalization performance compared to random selection. We also differentiate between semantic interpretation errors and performance loss, demonstrating that the latter is not solely determined by the former. Importantly, limitations in the encoding capabilities of a target model can be mitigated by employing a more powerful encoder at the transmitter.

For future work, we aim to optimize the similarity function, refine the equalization algorithm, and explore more effective clustering techniques for prototypical anchors to enhance robustness and efficiency.

\bibliographystyle{ieeetr}
\bibliography{bib}

\end{document}